%% file: main.tex
\documentclass[runningheads]{llncs}
\usepackage[T1]{fontenc}
\usepackage{amsmath}
\usepackage{amssymb}
\usepackage[hidelinks]{hyperref}
\usepackage{graphicx,verbatim}
\usepackage{booktabs}
\usepackage{multirow}
\usepackage{makecell}
\usepackage{threeparttable}
\usepackage{pifont}
\usepackage[table]{xcolor}
\newcommand{\cmark}{\ding{51}}
\newcommand{\corrauth}{\textsuperscript{\ensuremath{\dagger}}}
\newsavebox\CBox
\newcommand*\textBF[1]{\sbox\CBox{#1}\resizebox{\wd\CBox}{\ht\CBox}{\textbf{#1}}}

\begin{document}
\title{OTCHA: Optimal Transport-driven Confidence-aware Latent Hub Alignment for Multi-View Medical Image Classification}
\titlerunning{OTCHA: Optimal Transport-driven Confidence-aware Latent Hub Alignment for Multi-View Medical Image Classification}
%

\author{Jiwoong Yang\inst{1} \and
Haejun Chung\inst{1}\corrauth \and
Ikbeom Jang\inst{2}\corrauth}
\authorrunning{J. Yang et al.}
\institute{Hanyang University, Seoul, Republic of Korea \\
\email{\{jiwoong, haejun\}@hanyang.ac.kr} \and
Hankuk University of Foreign Studies, Yongin, Republic of Korea \\
\email{ijang@hufs.ac.kr}}
  
\maketitle              
\begingroup
\renewcommand{\thefootnote}{\ensuremath{\dagger}}
\footnotetext{Corresponding author.}
\endgroup
\input{sections/0_abstract}  
\input{sections/1_introduction}  
\input{sections/2_methods}  
\input{sections/3_experiments}  
\input{sections/4_conclusion}  
\subsubsection{\discintname} The authors have no competing interests.
%
%
\bibliographystyle{splncs04}
\bibliography{mybibliography}

\end{document}

%% file: sections/0_abstract.tex
\begin{abstract} 
Multi-view imaging, such as mammography and chest radiography, is a standard component of clinical practice. However, medical images are often unregistered and contain view-specific artifacts or irrelevant background cues that can obscure diagnostically relevant findings. Many existing methods directly fuse per-view representations, allowing such irrelevant content to contaminate the fused embedding and reducing robustness under varying view configurations.
We propose OTCHA, a confidence-aware latent hub token alignment module based on optimal transport (OT) that refines patch tokens before fusion for multi-view classification. OTCHA introduces a set of learnable latent hub tokens shared across views.
For each view, we compute an OT plan between patch tokens and hub tokens that jointly considers feature similarity and geometry, and augment the OT formulation with token-conditional dustbins to enable partial matching and discard irrelevant tokens. The resulting transport plan provides token-wise matching confidence, which gates hub-mediated message passing and weights a novel optimal-transport-based representation alignment loss to stabilize refinement. 
Experiments on three multi-view medical image datasets demonstrate consistent improvements over competing baselines across diverse anatomies and view configurations. Our code is available at \url{https://github.com/labhai/OTCHA}.

\keywords{Multi-view Medical Image Fusion \and Confidence-Aware Fusion \and Latent Hub \and Token-Level Refinement \and Cross-view Learning}

\end{abstract}

%% file: sections/1_introduction.tex
\section{Introduction}
\label{sec:intro}

While 3D medical imaging is increasingly used for complex diagnosis, 2D imaging, such as X-ray, 2D ultrasound, and mammography, remains dominant in most countries. However, a single 2D projection inevitably superimposes overlapping anatomical structures, limiting the visibility of clinically relevant findings. To mitigate this limitation, standard clinical practice involves acquiring multiple views; this enables cross-referencing of complementary information from distinct projections. With recent advances in deep learning, multi-view learning methods have been developed to jointly leverage these multiple projections for improved prediction~\cite{ji2023mammo,wan2024correlation,manigrasso2025mammography}. However, robust multi-view fusion remains challenging in real-world clinical scenarios. Clinical images are often unregistered and vary in coverage and acquisition; thus, many patch tokens correspond to background or view-specific artifacts. When dense tokens are fused without explicitly modeling token reliability, irrelevant content can contaminate the fused embedding and obscure localized discriminative cues.

Prior methods in multi-view medical image analysis often directly fuse per-view representations. Early approaches such as MVCNet~\cite{zhu2021mvc} employ late fusion by concatenating or pooling view-wise features. Subsequent approaches strengthen cross-view interaction via cross-view attention and the exchange of token features~\cite{van2021multi,sun2022transformer,sarker2024mv}. Recent methods enforce consistency between single-view and fused predictions via mutual distillation~\cite{black2024multi} and explore state-space sequence models for efficient cross-view exchange~\cite{yang2024cardiovascular,zheng2025xfmamba}. To better address unregistered views, several works have also begun to explicitly learn local cross-view correspondences to guide attention~\cite{du2025geometry,dabboussi2025self}. However, these methods offer limited control over token reliability and partial matching, allowing irrelevant background or view-specific artifacts to dilute clinically informative cues in the fused representation. Several methods are designed for domain-specific use, limiting scalability to datasets with varying numbers and compositions of views.
In this context, optimal transport (OT) offers a principled framework for learning regularized set-to-set correspondences, where the resulting transport plan can naturally derive confidence measures. 
OT has been widely used for patch- and feature-level correspondence~\cite{ni2023pats,le2024integrating} and robust matching with partial or unbalanced formulations~\cite{xu2024temporally,de2023unbalanced}, often incorporating rejection mechanisms such as dustbins~\cite{sarlin2020superglue,izquierdo2024optimal}. In medical imaging, OT-based approaches are also gaining traction, including cross-view representation learning~\cite{gorade2025otcxr} and multimodal alignment~\cite{shaaban2025motor}. However, using OT to provide token-level partial matching or token-wise confidence remains underexplored in multi-view medical image classification.

In this paper, we propose Optimal Transport Confidence-aware Latent Hub Alignment (OTCHA) for multi-view medical image classification.
Unlike conventional fusion designs, OTCHA introduces a set of learnable latent hub tokens as a shared meeting point across views, enabling token refinement prior to fusion. For each view, OTCHA computes an entropic transport plan that aligns patch tokens to hub tokens using a fused cost combining feature similarity with a geometry-aware prior, augmented with token-conditional dustbins for partial matching. The resulting transport plan provides token-level matching confidence, which gates hub-mediated message passing and weights a novel Optimal Transport Representation Alignment (OTRA) loss to stabilize refinement. The main contributions are as follows: 
(1) We introduce OTCHA, a module that refines per-view patch tokens via feature–geometry fused OT with token-conditional dustbins to a learnable latent hub before fusion.
(2) We propose OTRA, a confidence-weighted alignment loss with stop-gradient targets that stabilizes hub-mediated refinement.
(3) OTCHA is motivated by radiological cross-referencing practice, and its hub assignment map offers interpretable cross-view correspondences.
(4) Experiments on three multi-view medical image datasets across distinct anatomies show consistent improvements over state-of-the-art methods.

%% file: sections/2_methods.tex
\section{Method}
\label{sec:method}
\begin{figure}[t]
\centering
\includegraphics[width=\textwidth]{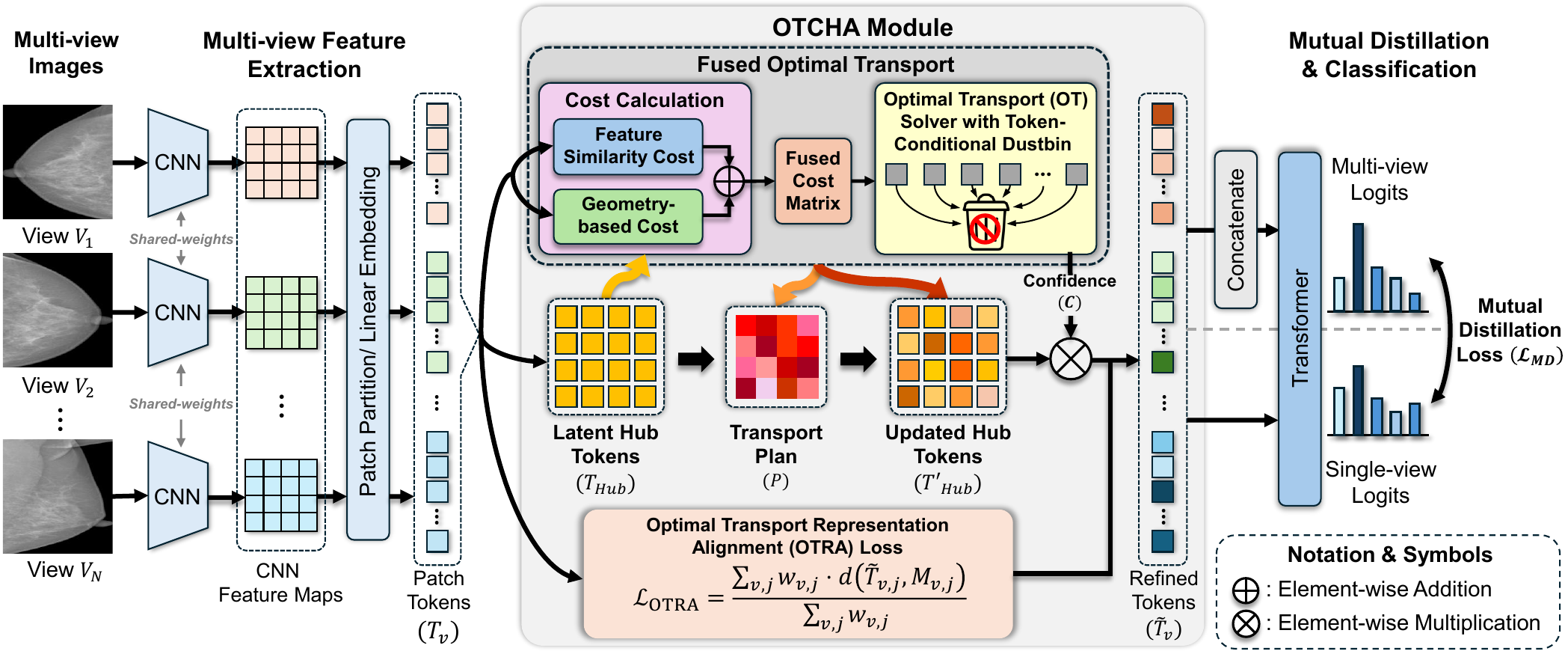}
\caption{Overview of the proposed framework. OTCHA is inserted between patch embedding and multi-view fusion. It refines per-view patch tokens before fusion using feature- and geometry-based optimal transport to a latent hub, where token-conditional dustbins reject unmatched or irrelevant tokens. OT-based matching confidence guides hub-mediated refinement and weights the OTRA alignment loss, and the refined tokens are fused for classification with mutual distillation.}
\label{fig:otcha_overview}
\end{figure}

Given $N$ multi-view images $\{\mathbf{I}_v\}_{v=1}^{N}$, we predict the clinical condition by leveraging complementary information across views. OTCHA refines per-view tokens before multi-view fusion as illustrated in Fig.~\ref{fig:otcha_overview}. It performs (i) confidence-aware OT matching to a learnable latent hub with token-conditional dustbins and (ii) hub-mediated message passing. We further regularize the refinement process using the proposed confidence-weighted OTRA loss.
\subsection{Preliminaries}
\label{sec:preliminaries}
\noindent\textbf{Base Framework.}
We adopt the shared CNN–Transformer encoder and mutual distillation framework of MV-HFMD~\cite{black2024multi} as the backbone for fair comparison. Each view image $\mathbf{I}_v$ is encoded by a shared CNN–Transformer to a patch-token sequence $\mathbf{X}_v \in \mathbb{R}^{S \times D}$, augmented with learnable positional and view-identity encodings. Multi-view prediction is obtained by concatenating tokens from all views and passing them through the Transformer, while single-view logits are computed in parallel for mutual distillation as in \cite{black2024multi}.
\\
\noindent\textbf{Optimal Transport.}
Optimal transport (OT) finds a minimum-cost assignment between two sets of elements. Given a source set of $S$ elements and a target set of $M$ elements, OT solves for a transport plan $\mathbf{P} \in \mathbb{R}_{\geq 0}^{S \times M}$ that minimizes the total cost $\langle \mathbf{C},\, \mathbf{P} \rangle$ subject to marginal constraints $\mathbf{P}\mathbf{1}_M = \boldsymbol{\mu}$ and $\mathbf{P}^\top\mathbf{1}_S = \boldsymbol{\nu}$, where $\mathbf{C} \in \mathbb{R}^{S \times M}$ is a pairwise cost matrix and $\boldsymbol{\mu} \in \mathbb{R}^{S}$, $\boldsymbol{\nu} \in \mathbb{R}^{M}$ are the source and target marginals. We use entropic OT solved by the Sinkhorn algorithm~\cite{cuturi2013sinkhorn} with regularization $\epsilon$; when using an affinity matrix $\mathbf{A}$, we set $\mathbf{C} = -\mathbf{A}$.
\subsection{Proposed Method}
\label{sec:otcha}
Our proposed method \textbf{OTCHA} introduces learnable latent hub tokens $\mathbf{H} \in \mathbb{R}^{M \times D}$ as a shared intermediary, where $M$ is set to match the number of spatial positions in the CNN feature map. For each view, an optimal transport plan aggregates patch-token information into the hub, and the hub then broadcasts the aggregated information back to the view, yielding refined tokens $\tilde{\mathbf{X}}_v$.
\\
\noindent\textbf{Fused Optimal Transport.}
For each view $v$, we compute a transport plan between raw patch tokens $\mathbf{X}_v = \{\mathbf{x}_{v,j}\}_{j=1}^{S}$ and latent hub tokens $\mathbf{H} = \{\mathbf{h}_k\}_{k=1}^{M}$ using a fused score matrix. We obtain normalized query and key representations $\mathbf{Q}_v \in \mathbb{R}^{S \times D}$ and $\mathbf{K}_h \in \mathbb{R}^{M \times D}$ via layer normalization and learned linear projections, and compute the fused affinity:
\begin{equation}
  \mathbf{A} = \mathbf{Q}_v \mathbf{K}_h^\top
             + \lambda_{\text{geo}} \cdot \mathbf{A}_{\text{geom}},
  \label{eq:fused_score}
\end{equation}
where $\mathbf{A}_{\text{geom}}(j,k) = -\|\mathbf{p}_{v,j} - \tanh(\mathbf{p}_{h,k})\|^2 / 2\sigma^2$ is a Gaussian kernel over learnable hub positions $\mathbf{p}_{h,k}$. Here, $\mathbf{p}_{v,j}$ are normalized 2D patch coordinates on the $S$--token grid, and $\sigma$ is a fixed bandwidth.
\\
\noindent\textbf{Token-Conditional Dustbin.}
Among all view tokens, some may contain irrelevant information such as background regions or imaging artifacts, which can introduce noise into the hub representation if forced to participate in transport. To enable partial matching that excludes such tokens, we augment the fused score matrix $\mathbf{A} \in \mathbb{R}^{S \times M}$ with a dustbin row and column:
\begin{equation}
  \bar{\mathbf{A}}
  = \begin{bmatrix}
      \mathbf{A}        & \mathbf{b}_u \\
      \mathbf{b}_h^\top & b_0
    \end{bmatrix}
  \!\in \mathbb{R}^{(S+1) \times (M+1)},
  \label{eq:augmented_score}
\end{equation}
where $b_{u,j} = b_0 + \mathrm{MLP}_{\text{bin}}(\mathrm{LN}(\mathbf{x}_{v,j}))$ and $b_{h,k} = b_0 + \mathrm{MLP}_{\text{bin}}(\mathrm{LN}(\mathbf{h}_k))$, with $b_0$ being a learnable base score. Unlike a global dustbin~\cite{sarlin2020superglue}, this token-conditional design lets each token independently control its unmatched mass based on its content. We solve for $\mathbf{P}^* \!\in\! \mathbb{R}^{(S+1) \times (M+1)}$ via the Sinkhorn algorithm~\cite{cuturi2013sinkhorn} with entropic regularization $\epsilon$ and rectangular marginals $\boldsymbol{\mu} = [\mathbf{1}_S;\, M]$, $\boldsymbol{\nu} = [\mathbf{1}_M;\, S]$. $\mathbf{1}_S$ and $\mathbf{1}_M$ denote all-ones vectors.
\\
\noindent\textbf{Matching Confidence.}
From $\mathbf{P}^*$, we derive confidence scores that modulate the subsequent message passing and representation alignment.
Let $\mathbf{P} = \mathbf{P}^*_{1:S,\,1:M}$ denote the transport plan excluding dustbin entries. The view-token confidence $c_{v,j} = 1 - \mathbf{P}^*_{j,\,M+1}$ measures how much mass token $j$ retains for matching rather than being discarded. The hub-token confidence $c^{(v)}_{h,k} = \sum_{j} \mathbf{P}^{(v)}_{j,k}$ reflects the total mass hub token $k$ receives from the view. The view-level reliability $\beta_v = \frac{1}{S}\sum_{j} c_{v,j}$ summarizes the overall matching quality of view $v$.
\\
\noindent\textbf{Hub-Mediated Message Passing.}
The hub collects information from all views and broadcasts the fused representation back to each view.

\noindent\textbf{(i) View\,$\rightarrow$\,Hub: }
For each view $v$, we compute a weighted message from view tokens to each hub token $k$ using the transport plan $\mathbf{P}$:
\begin{equation}
  \mathbf{m}_{h,k}
  = \frac{\sum_{j} \mathbf{P}_{j,k}\,
          \mathbf{W}_V \mathbf{x}_{v,j}}
         {c_{h,k}+\delta},
  \label{eq:hub_msg}
\end{equation}
where $\mathbf{W}_V$ is a learned value projection. Here and in the following, we add a small constant $\delta$ to denominators for numerical stability. A learned sigmoid gate $g_{h,k} = \sigma(\mathrm{MLP}_{\text{gate}}([\mathbf{h}_k;\,c_{h,k};\,\beta_v]))$ modulates the message based on the hub token and matching confidence, yielding the update $\Delta\mathbf{h}_{v,k} = g_{h,k} \cdot c_{h,k} \cdot \mathbf{m}_{h,k}$.
The hub is updated by averaging across all $N$ views
with a residual connection:
$\mathbf{H} \leftarrow \mathbf{H}
  + \mathbf{W}_O\!\big(\frac{1}{N}\sum_{v}
    \Delta\mathbf{H}_v\big)$,
where $\mathbf{W}_O$ is a learned output projection.

\noindent\textbf{(ii) Hub\,$\rightarrow$\,View: }
Since the hub now carries aggregated multi-view information, the correspondences differ from the initial step. Using the updated hub, we recompute the fused optimal transport (Eq.~\eqref{eq:fused_score}--\eqref{eq:augmented_score}) to obtain a new transport plan $\tilde{\mathbf{P}}$, confidence scores $\tilde{c}_{v,j}$, and gate $\tilde{g}_{v,j}$.
Each view token is then refined as:
\begin{equation}
  \tilde{\mathbf{x}}_{v,j}
  = \mathbf{x}_{v,j}
    + \mathbf{W}_O\!\Big(
        \tilde{g}_{v,j} \cdot \tilde{c}_{v,j} \cdot
        \frac{\sum_{k} \tilde{\mathbf{P}}_{j,k}\,
              \mathbf{W}_V \mathbf{h}_k}
             {\tilde{c}_{v,j}+\delta}
      \Big).
  \label{eq:hub_to_view}
\end{equation}
All projection weights are shared across both message-passing directions for parameter efficiency.
\subsection{Loss Function}
\label{sec:otra_loss}
We train OTCHA with the same classification and mutual distillation loss used in \cite{black2024multi}, and add our OTRA loss to stabilize hub-mediated refinement. The overall training objective is:
\begin{equation}
  \mathcal{L}
  = \mathcal{L}_{\text{s}} + \mathcal{L}_{\text{m}}
  + \alpha\,\mathcal{L}_{\text{MD}}
  + \gamma\,\mathcal{L}_{\text{OTRA}},
  \label{eq:total_loss}
\end{equation}
where $\mathcal{L}_{\text{s}}$ and $\mathcal{L}_{\text{m}}$ denote the single-view and multi-view classification losses, respectively, and $\mathcal{L}_{\text{MD}}$ is the mutual distillation loss defined in base framework~\cite{black2024multi}. We keep $\alpha$ identical to the setting in \cite{black2024multi}, and $\gamma$ weights the OTRA term.

\noindent\textbf{OTRA Loss.}
OTRA provides an explicit token-level learning signal for the hub-mediated message passing and stabilizes training. For each view token $\tilde{\mathbf{x}}_{v,j}$, we define its hub-to-view message $\tilde{\mathbf{m}}_{v,j} = \sum_{k}\tilde{\mathbf{P}}_{j,k}\,\mathbf{W}_V\mathbf{h}_k\,/\,(\tilde{c}_{v,j}+\delta)$ in Eq.~\eqref{eq:hub_to_view}, and measure a cosine distance after $\ell_2$-normalization ($\hat{\mathbf{x}} = \mathrm{Norm}_{\ell_2}(\mathrm{LN}(\mathbf{x}))$), weighted by the learned gate and OT-derived view reliability:
\begin{equation}
  \mathcal{L}_{\text{OTRA}}
  = \frac{
      \sum_{v,j}\;
        \overbrace{\tilde{g}_{v,j}\;\beta_v}
                  ^{w_{v,j}}
        \;\Big(1
          - \hat{\tilde{\mathbf{x}}}_{v,j}^{\top}\,
            \hat{\tilde{\mathbf{m}}}_{v,j}
        \Big)
    }{
      \sum_{v,j}\, w_{v,j} +\delta
    },
  \label{eq:otra_loss}
\end{equation}
where $w_{v,j}$ combines the gate $\tilde{g}_{v,j}$ and the OT-derived view reliability $\beta_v$ (Sec.~\ref{sec:otcha}).  We stop gradients through the hub message and the confidence weights to prevent degenerate solutions and stabilize optimization.

%% file: sections/3_experiments.tex
\section{Experiments and Results}
\label{sec:experiments}
\subsection{Experimental Details}
\noindent\textbf{Datasets and Evaluation.} We evaluate OTCHA on three public multi-view datasets. For all datasets, we perform patient-level splits into train/validation/test splits to prevent data leakage.
The VinDr-Mammo~\cite{nguyen2023vindr} is a full-field digital mammography dataset labeled with BI-RADS categories. We formulate a binary task by mapping BI-RADS 1--2 to normal and 3--5 to abnormal and following~\cite{ghosh2024mammo} for ROI cropping and background suppression. Each study contains four standard views (LCC, LMLO, RCC, RMLO). We evaluate (i) a breast-level setting using two views per breast and (ii) a study-level setting using all four views, where a study is labeled abnormal if at least one breast is abnormal. The breast-level splits contain 7{,}199/800/2{,}000 samples and the study-level splits contain 3{,}599/400/1{,}000 studies.
The MURA~\cite{rajpurkar2017mura} is a musculoskeletal abnormality detection dataset formulated as a binary classification. We select studies with at least two radiographic views and randomly sample two per study, yielding 11{,}056/1{,}270/1{,}098 studies.
The CheXpert~\cite{irvin2019chexpert} contains chest radiographs annotated for 13 distinct observations. We use VisualCheXbert labels~\cite{jain2021visualchexbert} and include only studies with both frontal and lateral views, and split them into 23{,}577/3{,}923/3{,}913 studies. For all datasets, images are resized to 224$\times$224. We report AUROC for all datasets, using the average over 13 labels for CheXpert. All results are reported as mean and standard deviation over five training runs.
\\
\noindent\textbf{Implementation Details.} OTCHA is implemented in PyTorch and trained on an NVIDIA RTX 6000 Ada GPU. The backbone is a shared hybrid CNN--Transformer (ResNet26$+$ViT small pre-trained on ImageNet~\cite{deng2009imagenet}). We set $\alpha=0.1$ following \cite{black2024multi}. Key hyperparameters were selected via grid search, yielding $\lambda_{\text{geo}}=0.2$, entropic regularization $\epsilon=0.2$ with Sinkhorn for 20 iterations. We employ $\gamma=0.3$ in Eq.~\eqref{eq:total_loss}.
We use a batch size of 32, a learning rate of $1e^{-3}$ with SGD, weight decay of $5e^{-4}$, and a cosine annealing scheduler (min LR $1e^{-6}$). The model is trained for up to 100 epochs with early stopping.

\subsection{Results and Analysis}
\input{tables/table_1}
\subsubsection{Comparison with State-of-the-Arts.} We compare our method against SOTA multi-view methods across three medical imaging datasets, as summarized in Table~\ref{tab:comparison_full}. In all datasets and view configurations, OTCHA achieves the best performance with only marginal parameter overhead. This demonstrates that refinement-before-fusion can substantially improve multi-view recognition without sacrificing efficiency. Notably, the performance gain is most pronounced in the study-level (4-view) setting of VinDr-Mammo, suggesting that OTCHA offers significant benefits when integrating multiple complementary views with richer information. Overall, OTCHA consistently outperforms prior methods, achieving statistically significant improvements over most baselines, validating its effectiveness across diverse anatomies and view configurations.
\input{tables/table_2}

\begin{figure}[t]
\centering
\includegraphics[width=\textwidth]{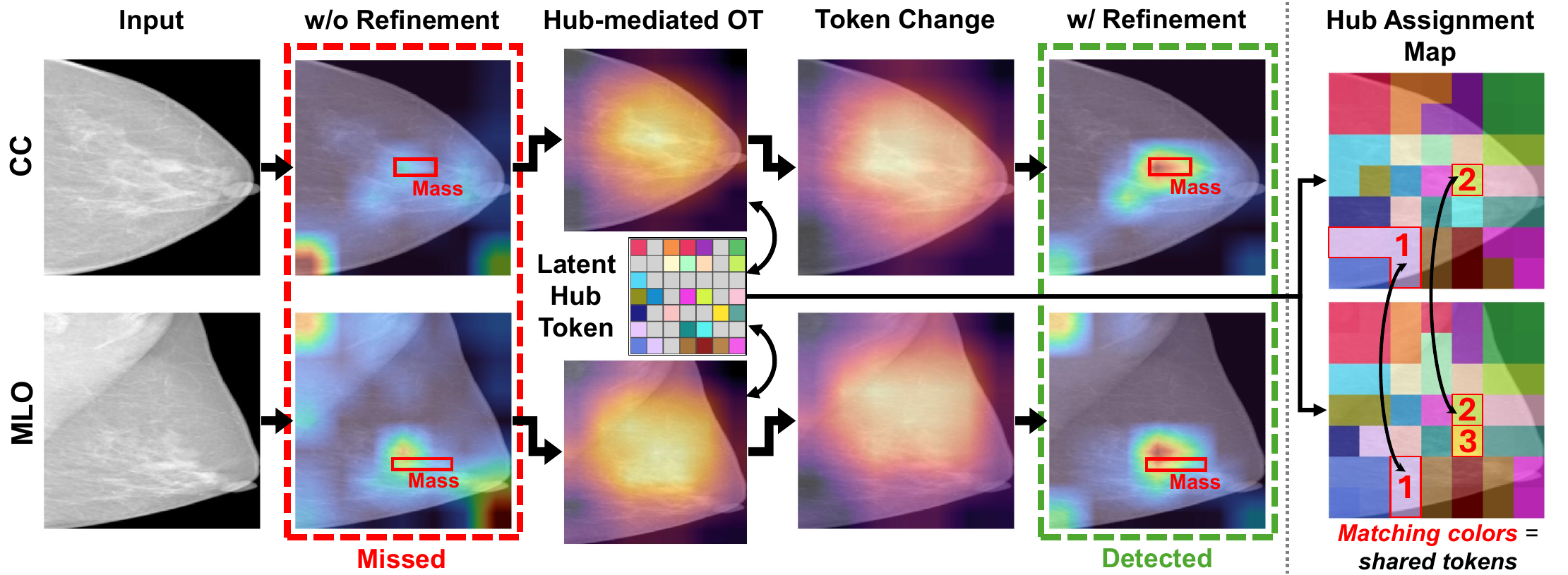}
\caption{Qualitative visualization of the refinement process. The hub assignment map colors each patch by its assigned hub; matching colors across views indicate cross-view correspondence, while view-specific colors represent regions without counterparts in the other view.}
\label{fig:otcha_visualization}
\end{figure}

\begin{figure}
\centering
\includegraphics[width=\textwidth]{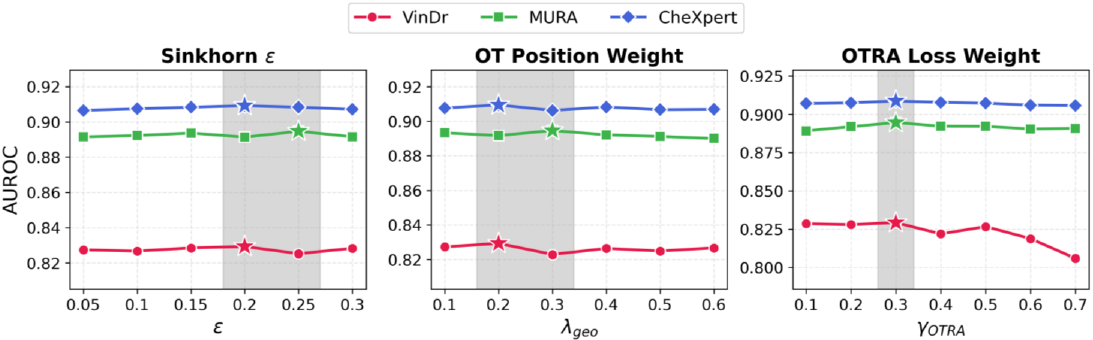}
\caption{Sensitivity analysis of key hyperparameters. ($\bigstar=$ best result)}
\label{fig:hyperparam_sens}
\end{figure}

\noindent\textbf{Ablation studies (Table~\ref{tab:ablation}).} \textbf{(I):} Baseline without OTCHA and OTRA. \textbf{(II):} Add the OTCHA module with fused OT and token-conditional dustbins, but without OTRA loss. \textbf{(III):} Incorporate OTRA loss with uniform weighting. \textbf{(IV):} Remove dustbins entirely, reducing to balanced OT. \textbf{(V):} Replace token-conditional dustbins with a global dustbin. \textbf{(VI):} Remove the geometry cost from fused OT.  \textbf{(VII):} Full model with all components and confidence-weighted OTRA. (II)–(III) and (VII) show that each component incrementally contributes to the final performance. (IV)–(VI) further confirm that geometry cost, partial matching, and token-level dustbin control each provide complementary gains.
\\
\noindent\textbf{Qualitative evaluation (Fig.~\ref{fig:otcha_visualization}).}
We visualize the OTCHA process on a representative mammography case. Without refinement, the model attends broadly to background regions, producing weak activations around the mass. After hub-mediated OT, background activations are suppressed while the mass region receives substantially stronger responses, resulting in localized and concentrated attention. The hub assignment map confirms semantically meaningful token-level correspondences between CC and MLO views. By suppressing view-specific artifacts and reinforcing shared findings across views, OTCHA reflects how radiologists cross-reference views to confirm suspicious regions.
\\
\noindent\textbf{Hyperparameter sensitivity (Fig.~\ref{fig:hyperparam_sens}).}
We analyze the sensitivity of Sinkhorn regularization ($\epsilon$), geometry cost weight ($\lambda_\text{geo}$), and OTRA loss weight ($\gamma$) across all datasets. Performance remains stable over a wide range of $\epsilon$ and $\lambda_\text{geo}$, demonstrating robustness to these hyperparameters. When $\gamma=0.3$, optimal performance is achieved, while larger values degrade performance as the alignment term dominates the classification objective.

%% file: tables/table_1.tex
\begin{table}[t]
\centering
\caption{OTCHA achieves the best performance on all dataset configurations. `$*$': statistically significant (Mann-Whitney U, FDR-corrected $p < 0.05$). `--': not supported.}
\label{tab:comparison_full}
\newcommand{\pd}{\phantom{$^{\dagger}$}}
\setlength{\tabcolsep}{2.0pt}
\fontsize{8}{10}\selectfont
\begin{tabular}{@{}lccccccc@{}}
\toprule
\multirow{2}{*}{Method} & \multirow{2}{*}{Backbone} & \multirow{2}{*}{\#P(M)} & \multicolumn{2}{c}{VinDr-Mammo} & \multirow{2}{*}{MURA} & \multirow{2}{*}{CheXpert}\\
\cmidrule(lr){4-5}
 & & & breast-level & study-level & &  \\
\midrule
MVC-Net~\cite{zhu2021mvc}   & ResNet26     & 71.10 & 0.783$\pm$.007$^{*}$ & -- & 0.859$\pm$.003$^{*}$ & 0.900$\pm$.001$^{*}$ \\
CVT~\cite{van2021multi}       & ResNet26     & 63.55 & 0.786$\pm$.009$^{*}$ & -- & 0.849$\pm$.005$^{*}$ & 0.898$\pm$.002$^{*}$ \\
MVT~\cite{chen2021mvt}       & DeiT-s       & 21.67 & 0.782$\pm$.013$^{*}$ & 0.748$\pm$.012$^{*}$ & 0.852$\pm$.006$^{*}$ & 0.896$\pm$.002$^{*}$  \\
ETMC~\cite{han2022trusted}      & ResNet26     & 22.35 & 0.797$\pm$.007$^{*}$ & 0.745$\pm$.008$^{*}$ & 0.869$\pm$.005$^{*}$ & 0.903$\pm$.001$^{*}$  \\
MV-Swin-T~\cite{sarker2024mv} & Swin-T       & 58.32 & 0.615$\pm$.008$^{*}$ & -- & 0.704$\pm$.006$^{*}$ & 0.878$\pm$.006$^{*}$  \\
MV-HFMD~\cite{black2024multi}   & ViT-s        & 36.05 & 0.813$\pm$.008\pd & 0.759$\pm$.003$^{*}$ & 0.886$\pm$.003$^{*}$ & 0.904$\pm$.002$^{*}$ \\
XFMamba~\cite{zheng2025xfmamba}   & VMamba-s     & 58.73 & 0.816$\pm$.004$^{*}$ & -- & 0.889$\pm$.002$^{*}$ & 0.906$\pm$.001$^{*}$ \\
\hline
\rowcolor{gray!12}
OTCHA     & ViT-s        & 36.73 & \textBF{0.826$\pm$.006\pd} & \textBF{0.782$\pm$.005\pd} & \textBF{0.895$\pm$.003\pd} & \textBF{0.908$\pm$.002\pd} \\
\bottomrule
\end{tabular}
\end{table}

%% file: tables/table_2.tex
\begin{table}[t]
\centering
\caption{Ablation studies on VinDr-Mammo (study-level) demonstrate the significant improvement of the baseline.}
\label{tab:ablation}
\setlength{\tabcolsep}{4pt}
\fontsize{8}{10}\selectfont
\begin{tabular}{clccccccc}
\toprule
\multirow{2}{*}{Num.} 
& \multirow{2}{*}{Variant} 
& \multicolumn{3}{c}{\textBF{OTCHA module}} 
& \multicolumn{2}{c}{\textBF{OTRA}} 
& \multirow{2}{*}{AUROC $\uparrow$}
& \multirow{2}{*}{$\Delta$ (\%)}\\
\cmidrule(lr){3-5}\cmidrule(lr){6-7}
& & Geo & Dust. & T-cond. & Loss & Conf-wt. & \\
\midrule
(I) & Baseline
&  &  &  &  &  & 0.7589 & - \\
(II) & $+$ OTCHA
& \cmark & \cmark & \cmark &  &  & 0.7628 & \textcolor{red}{$+0.39$} \\
(III) & $+$ OTRA loss
& \cmark & \cmark & \cmark & \cmark &  & 0.7759 & \textcolor{red}{$+1.70$} \\
\hline
(IV) & w/o dustbin
& \cmark &  &  & \cmark &  & 0.7731 & \textcolor{red}{$+1.42$} \\
(V) & global dustbin
& \cmark & \cmark &  & \cmark & \cmark & 0.7747 & \textcolor{red}{$+1.58$} \\
(VI) & w/o geometry
&  & \cmark & \cmark & \cmark & \cmark & 0.7763 & \textcolor{red}{$+1.74$} \\
\hline
\rowcolor{gray!12}
(VII) & \textBF{Ours} 
& \cmark & \cmark & \cmark & \cmark & \cmark & \textBF{0.7816} & \textcolor{red}{$+$\textBF{2.27}} \\
\bottomrule
\end{tabular}
\end{table}

%% file: sections/4_conclusion.tex
\section{Conclusion}
\label{sec:conclusion}
We propose OTCHA, a confidence-aware refinement-before-fusion module that leverages optimal transport with token-conditional dustbins. It refines per-view patch tokens via hub-mediated message passing. A confidence-weighted OTRA loss further stabilizes refinement by aligning refined tokens with their hub-broadcast messages. 
By suppressing clinically irrelevant image tokens, OTCHA aligns with how radiologists selectively attend to and cross-reference clinically meaningful regions across views. Moreover, the hub assignment map provides interpretable token-level correspondences across views, potentially facilitating the localization and verification of suspicious findings. Experiments across three datasets/anatomies show consistent performance gains with marginal parameter overhead, suggesting a promising direction for diverse multi-view medical applications.
The work is limited to fixed-view settings; future work will explore missing-view scenarios and additional modalities and downstream tasks.